\documentclass{article}

\usepackage{microtype}
\usepackage{graphicx}
\usepackage{subfigure}
\usepackage{booktabs} 
\usepackage{amssymb}
\usepackage{amsmath}
\usepackage{amsthm}
\usepackage{amsfonts}
\usepackage{dsfont}
\usepackage[utf8]{inputenc} 
\usepackage[T1]{fontenc}

\usepackage{bbm}
\usepackage{multirow}

\usepackage{tabularx}
\usepackage{arydshln}
\usepackage[hang,flushmargin]{footmisc}

\newcommand{\ignorethis}[1]{}

\newtheorem*{conjecture*}{Conjecture}

\newcommand{\comm}[1]{}

\newcommand\blfootnote[1]{%
  \begingroup
  \renewcommand\thefootnote{}\footnote{#1}%
  \addtocounter{footnote}{-1}%
  \endgroup
}

\makeatletter
\def\blfootnote{\gdef\@thefnmark{}\@footnotetext}
\makeatother

\usepackage{hyperref}
\usepackage{caption}
\usepackage{xurl}

\usepackage[accepted]{icml2021}

\icmltitlerunning{ Explaining Representation by Mutual Information }

\begin{document}

\twocolumn[
\icmltitle{ Explaining Representation by Mutual Information}


\vspace{0.2cm}
\centerline{\textbf{Lifeng Gu}} 
\centerline{Tianjin University}
\vskip 0.25in
]




\begin{abstract}
As interpretability gains attention in machine learning, there is a growing need for reliable models that fully explain representation content. We propose a mutual information (MI)-based method that decomposes neural network representations into three exhaustive components: total mutual information, decision-related information, and redundant information. This theoretically complete framework captures the entire input-representation relationship, surpassing partial explanations like those from Grad-CAM. Using two lightweight modules integrated into architectures such as CNNs and Transformers,we estimate these components and demonstrate their interpretive power through visualizations on ResNet and prototype network applied to image classification and few-shot learning tasks. Our approach is distinguished by three key features:

\begin{enumerate}
\item Rooted in mutual information theory, it delivers a thorough and theoretically grounded interpretation, surpassing the scope of existing interpretability methods.
\item Unlike conventional methods that focus on explaining decisions, our approach centers on interpreting representations.
\item It seamlessly integrates into pre-existing network architectures, requiring only fine-tuning of the inserted modules.
\end{enumerate}
\end{abstract}

\section{Introduction}
Representation learning has progressed rapidly~\cite{chen2020simple, grill2020bootstrap}, yet the essence of an effective representation remains unclear. Deep neural networks generate opaque representations, masking the connection between inputs and outputs. Methods like Grad-CAM~\cite{selvaraju2017gradcam} identify decision-relevant features but fail to capture the complete information encoded by local inputs. We propose a mutual information (MI)-based framework that decomposes the information between local inputs and representations into three exhaustive components: total mutual information \( I(Z,X_i) \), decision-related information \( I(Z, X_i') \), and redundant information \( R(Z, X_i) = I(Z, X_i) - I(Z, X_i') \). This approach fully accounts for \( I(X_i, Z) \), providing a holistic view of representation content. Using two lightweight modules, we estimate these components and validate their effectiveness through visualizations on ResNet and a prototype network.
\blfootnote{\textbf{Correspondence} \hfill Lifeng Gu — \texttt{gulifeng666@163.com}}
\blfootnote{\textbf{Code} \hfill \href{https://github.com/lifeng6666666/mire}{\texttt{github.com/lifeng6666666/mire}}}

\section{Related Works}

\subsection{Post-hoc Explanation Methods}
Post-hoc explanation methods aim to interpret neural network behavior by analyzing global aspects of their predictions, typically falling into two broad categories: visualization techniques for convolutional networks and attribution-based approaches that link network outputs or decisions to input features. Visualization methods emphasize intuitive depictions of network activity, whereas attribution-based techniques align more closely with our objective of providing precise, actionable insights into model behavior. In this section, we focus on attribution-based methods.
Gradient-based approaches, such as Gradient Maps~\cite{baehrens2010explain} and Saliency Maps~\cite{simonyan2013saliency}, compute the gradient of the output with respect to input features to identify regions driving network decisions. While straightforward and computationally efficient, their gradients are often unstable and sensitive to noise, reducing their trustworthiness. To overcome these weaknesses, Integrated Gradients~\cite{sundararajan2017axiomatic} averages gradients across multiple inputs, improving stability and mitigating noise. Other techniques, such as Layer-wise Relevance Propagation (LRP)~\cite{bach2015lrp}, Deep Taylor Decomposition (DTD)~\cite{montavon2017dtd}, and Grad-CAM~\cite{selvaraju2017gradcam}, propagate relevance scores backward through intermediate layers to assign importance to features, frequently relying on gradient-derived weights. A hybrid method, Guided Grad-CAM, integrates Guided Backpropagation with Grad-CAM to enhance stability and deliver more reliable explanations. Despite these improvements, post-hoc methods often fall short of revealing the underlying structure of learned representations---a limitation our approach addresses with a more comprehensive and theoretically grounded framework.

\subsection{Disentangled Representation Learning}
Disentangled representation learning focuses on interpreting neural networks by decomposing representations into individual components, assuming that data variations arise from independent generative factors---such as image orientation, lighting conditions, or object-specific attributes (e.g., hair length in portraits). The guiding principle is that a disentangled representation isolates these factors, such that modifying one component does not impact others. This characteristic is believed to facilitate downstream tasks; for example, in gender classification, a model could rely exclusively on the "gender" component of the representation.

A leading approach in this domain is the Variational Autoencoder (VAE), which enforces constraints on the posterior approximation \( z \sim q(z|x) \). The standard VAE objective balances reconstruction fidelity with regularization:
\[
L_{VAE} = \mathbb{E}_{q(z|x)}[\log p(x|z)] - D_{KL}(q(z|x) \| p(z)),
\]
where \( p(z) \) is typically a standard Gaussian prior. To enhance disentanglement, Higgins et al.~\cite{higgins2017beta} introduced \(\beta\)-VAE, which increases the weight of the KL divergence term:
\[
L_{\beta\text{-}VAE} = \mathbb{E}_{q(z|x)}[\log p(x|z)] - \beta D_{KL}(q(z|x) \| p(z)),
\]
where \( \beta > 1 \) encourages \( q(z|x) \) to match a prior with independent dimensions. Extending this, Kim and Mnih~\cite{kim2018disentangling} proposed FactorVAE, incorporating a total correlation (TC) penalty to further minimize dependencies among latent variables:
\[
L_{\text{FactorVAE}} = L_{VAE} + \lambda TC(q(z)),\]
where \( TC(q(z)) \) measures mutual dependencies and requires density ratio estimation for computation. While these methods excel in controlled settings, their reliance on the assumption of factor independence often proves restrictive in complex, real-world applications. In contrast, our approach harnesses mutual information to offer a more flexible and robust framework for interpreting representations, bypassing the constraints of strict independence assumptions.
\section{Method}

\subsection{Mutual Information Estimation}
Neural network representations encode information extracted from input data. To investigate this encoded information, we apply mutual information theory to quantify the mutual information between the representation \( f(X) \) and the input \( X \), denoted \( I(X, f(X)) \). This metric reflects how much input information is preserved in the representation. However, \( I(X, f(X)) \) alone offers limited insight into the specific details captured. To address this, we compute the mutual information between local input components \( X_i \) (e.g., pixels in an image or words in a sentence) and \( f(X) \), pinpointing the exact input elements encoded in the representation. This is formally defined as:

\begin{align}
I(X_i, f(X)) &= \mathbb{E}_{X_i} \left[ KL(p(f(X) \mid X_i) \parallel p(f(X))) \right] \notag \\
&= \mathbb{E}_{X_i} \left[ KL(p(X_i \mid f(X)) \parallel p(X_i)) \right],
\label{eq:mi}
\end{align}

where \( KL \) denotes the Kullback-Leibler divergence. Direct computation of the marginal distributions \( p(X_i) \) and \( p(f(X)) \), or the conditional distributions \( p(X_i \mid f(X)) \) and \( p(f(X) \mid X_i) \) in Equation~\eqref{eq:mi}, is computationally infeasible due to their complexity. To tackle this, we adopt the InfoNCE method~\cite{tschannen2019mutual} to estimate \( I(X_i, f(X)) \). Letting \( Z = f(X) \) represent the representation, the mutual information satisfies:

\begin{align}
I(X_i, f(X)) &= I(f(X), X_i) = I(Z, X_i) \notag \\
&\geq \mathbb{E}_{x_i \sim p(X_i), z \sim p(Z)} \left[ \log \frac{e^{f(z, x_i)}}{\frac{1}{N K} V} \right], \notag \\
V &= \sum_{n=1}^{N} \sum_{k=1}^{K} e^{f(z^{(n)}, x_k^{(n)})},
\label{eq:mi_lower}
\end{align}

where \( z^{(n)} \) is the representation of the \( n \)-th sample, and \( x_k^{(n)} \) is its \( k \)-th local component. Here, \( N \) denotes the batch size, and \( K \) is the number of local components per sample (e.g., pixels or words). We optimize this lower bound by maximizing it with respect to the parameters \( \theta \) of the Infomax estimator module, yielding:

\begin{equation}
I(Z, X_i) = \max_{\theta} \mathbb{E}_{x_i \sim p(X_i), z \sim p(Z)} \left[ \log \frac{e^{f(z, x_i)}}{\frac{1}{N K} V} \right].
\label{eq:max_mi_lower}
\end{equation}

The scoring function \( f(z, x_i) \) is computed by the Infomax estimator module, typically a single- or two-layer multilayer perceptron (MLP), assessing similarity between the representation and local inputs. This is illustrated in Figure~\ref{fig:infor_architecture}. While alternatives exist~\cite{hjelm2018learning, belghazi2018mine}, InfoNCE is chosen for its low variance and proven efficacy in representation learning.

\subsection{Information Bottleneck}
Our next goal is to extract decision-related information from the representation—information vital for preserving model decisions. We adopt the information bottleneck principle~\cite{tishby2015deep}, where \( Y \) denotes the label. This approach maximizes the mutual information between the representation \( Z \) and \( Y \), while minimizing the mutual information between \( Z \) and the input \( X \), formulated as:

\begin{equation}
\max I(Z, Y) \quad \text{subject to} \quad I(X, Z) \leq c
\label{eq:ib}
\end{equation}

Optimizing Equation~\eqref{eq:ib} directly is challenging due to its complexity. Inspired by masking mechanisms, we introduce a mask layer after the input \( X \), transforming each local component as \( x_i = x_i \cdot \lambda_i \), where \( \lambda_i \in [0, 1] \) controls the information flow for the \( i \)-th component. The mask is generated by a simple two-layer MLP or convolutional layer, which processes the input to produce \( \lambda_i \), then multiplies it with \( X \) before forwarding it to subsequent layers, as shown in Figure~\ref{fig:infor_architecture}. The optimization objective becomes:

\begin{equation}
\max_{\phi} \left[ l(X) - \beta \sum_i \lambda_i \right]
\label{eq:mi_min}
\end{equation}

Here, \( l(X) \) is the original objective function designed to maximize \( I(Z, Y) \), aligning the representation with the label. The term \( \beta \sum_i \lambda_i \) minimizes \( I(X, Z) \), with \( \beta \) as a hyperparameter balancing these goals. The parameters \( \phi \) govern the mask module, approximating the information bottleneck by retaining decision-critical information while suppressing excess details.

\subsection{Information Redundancy}
Finally, we aim to isolate redundant information within the representation—information irrelevant to decision-making and removable without affecting the model’s output. We first calculate \( I(X, Z) \), the mutual information between the input \( X \) and the representation \( Z \), and \( I(X', Z) \), the decision-related information, where \( X' \) is the masked input retaining only critical components. The redundancy for a local component \( X_i \) is:

\begin{equation}
R(Z, X_i) = I(Z, X_i) - I(Z, X_i')
\label{eq:redundant}
\end{equation}

Both \( I(Z, X_i) \) and \( I(Z, X_i') \) are estimated using the InfoNCE method from Equation~\eqref{eq:mi_lower}. Specifically, \( I(Z, X_i) \) measures the total information encoded from \( X_i \) into \( Z \), while \( I(Z, X_i') \) captures the decision-relevant portion after masking. Redundancy is derived by their difference.

To analyze the representation comprehensively using these three information types—mutual information, decision-related information, and redundant information—we integrate them into a unified objective function. Combining Equations~\eqref{eq:max_mi_lower} and \eqref{eq:mi_min}, the total objective is:

\begin{align}
L(X) = \max_{\theta, \phi} \Bigg[ l(X) &+ \alpha \mathbb{E}_{x_i \sim p(X_i), z \sim p(Z)} \left[ \log \frac{e^{f(z, x_i)}}{\frac{1}{N K} V} \right] \notag \\
&+ \alpha \mathbb{E}_{x_i' \sim p(X_i'), z \sim p(Z)} \left[ \log \frac{e^{f(z, x_i')}}{\frac{1}{N K} V'} \right] \notag \\
&- \beta \sum_i \lambda_i \Bigg],
\end{align}
\begin{align}
V &= \sum_{n=1}^{N} \sum_{k=1}^{K} e^{f(z^{(n)}, x_k^{(n)})}, \notag \\
V' &= \sum_{n=1}^{N} \sum_{k=1}^{K} e^{f(z^{(n)}, x_k'^{(n)})}
\label{eq:all}
\end{align}

\begin{figure*}[htbp]
    \centering
    \includegraphics[width=5.5in]{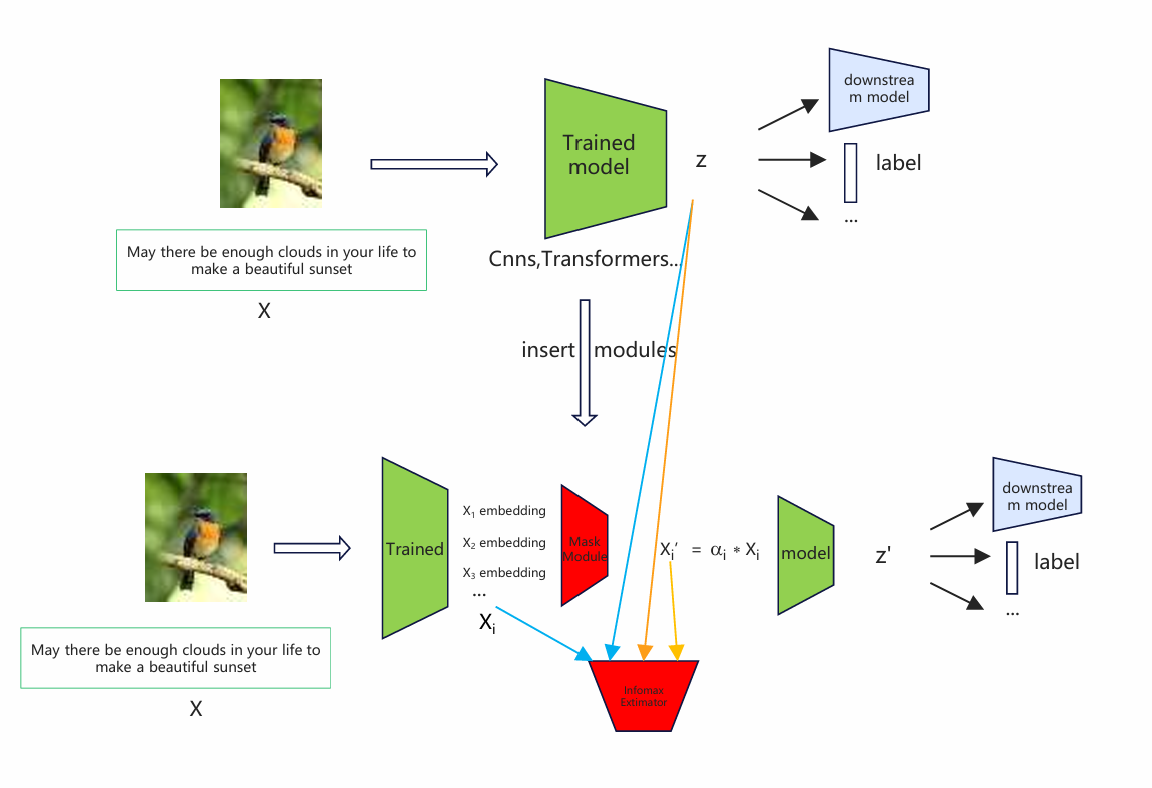}
    \caption{module architecture: the Infomax estimator processes local inputs $X_i$ and representation $Z$ to estimate $I(Z, X_i)$, while the mask module generates $\lambda_i$ to filter $X$ into $X'$, enabling $I(Z, X_i')$ computation.}
    \label{fig:infor_architecture}
\end{figure*}

In this expression, \( l(X) \) is the original objective function aimed at maximizing \( I(Z, Y) \), with \( \theta \) as the parameters of the Infomax estimator module and \( \phi \) as the parameters of the mask module. The first expectation term estimates \( I(Z, X_i) \), reflecting the mutual information between \( Z \) and \( X_i \). The second term, a novel addition, estimates \( I(Z, X_i') \) using the masked input \( X_i' \). Hyperparameters \( \alpha \) and \( \beta \) balance mutual information estimation and information bottleneck regularization, respectively. The terms \( V \) and \( V' \) are normalization factors for the unmasked and masked inputs. By optimizing Equation~\eqref{eq:all} via the Infomax and mask modules, we effectively estimate these three information types— \( I(Z, X_i) \),  \( I(Z, X_i') \) and \( R(Z, X_i) \),—without altering the original network’s parameters, requiring only fine-tuning of the two lightweight modules. The architecture is depicted in Figure~\ref{fig:infor_architecture}.

\section{Experiments}
Unlike other interpretability methods~\cite{baehrens2010explain, bach2015pixel, schulz2020restricting}, which are generally developed to explain model decisions, our approach centers on analyzing and interpreting model representations. In this section, we present experiments conducted on the ImageNet-Mini (a subset of 100 classes, 50,000 images) and CUB-200-2011 (200 bird species, 11,788 images). The Infomax estimator is a two-layer MLP with 256 hidden units, and the mask module is a two-layer CNN with $3\times3$ kernels. Hyperparameters are set as $\alpha = 1.0$, $\beta = 0.5$ \par
We select the output of an early layer in the network as the local representation of input samples and the output of a later layer as the global representation. By computing the mutual information between these two layers' outputs, as well as the mutual information between the masked output of the former and the global representation, we derive the three types of information targeted in our study: total information, decision-related information, and redundant information.

\subsection{Image Classification Visualization}
\begin{figure*}[htbp]
    \centering
    \includegraphics[width=5.5in]{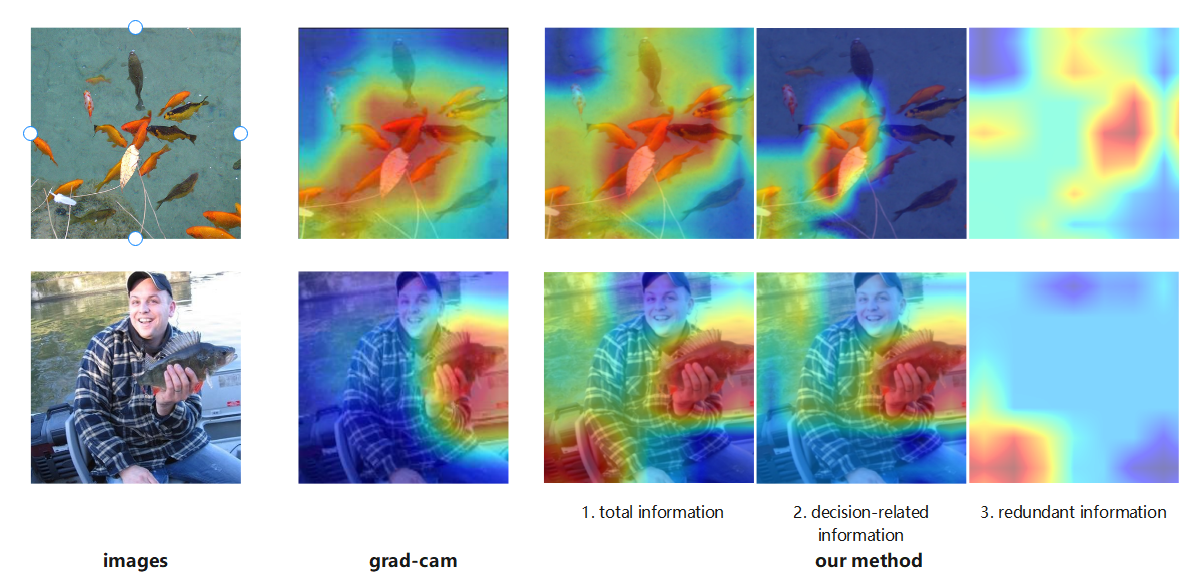}
    \caption{heatmap examples}
    \label{fig:visualization}
\end{figure*}

Image classification visualization is a benchmark task in interpretability research. We visualize images and their corresponding information heatmaps from the ImageNet-Mini dataset and compare our results with those of Grad-CAM. Using the ResNet50 model, we designate the intermediate convolutional layer of the Layer4 block as the local representation and the output of the final AvgPool layer as the global representation.
As shown in Figure\ref{fig:visualization}, our approach decomposes the model’s encoded information into three distinct types: total information, decision-related information, and redundant information. This provides a more comprehensive explanation than Grad-CAM. Specifically, in the visualization of the first image, the heatmap for total information closely resembles that of Grad-CAM. In the second image, the heatmap for decision-related information aligns similarly with Grad-CAM, highlighting the nuanced insights our method offers by separating these information components.
\subsection{Prototype Network Visualization}
\begin{figure*}[htbp]\centering                  
\subfigure[]{                    
\begin{minipage}{0.3\linewidth}\centering                                                          
\includegraphics[scale=0.3]{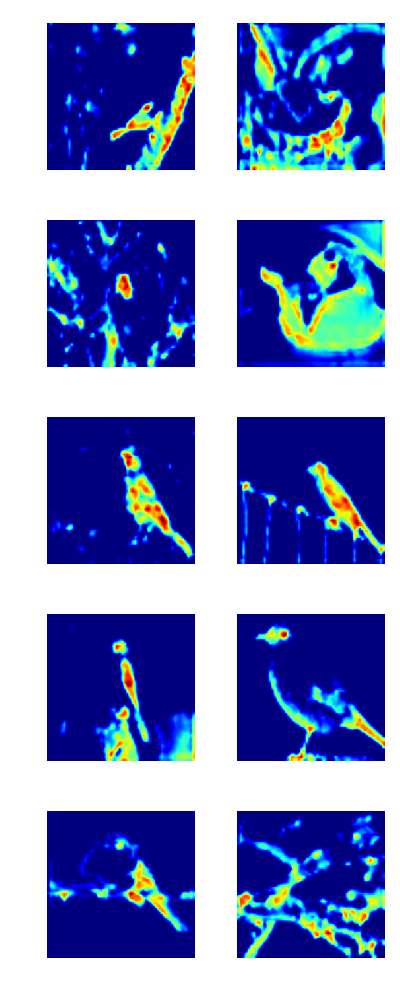}               
\end{minipage}}
\subfigure[]{                    
\begin{minipage}{0.3\linewidth}\centering                 
\includegraphics[scale=0.3]{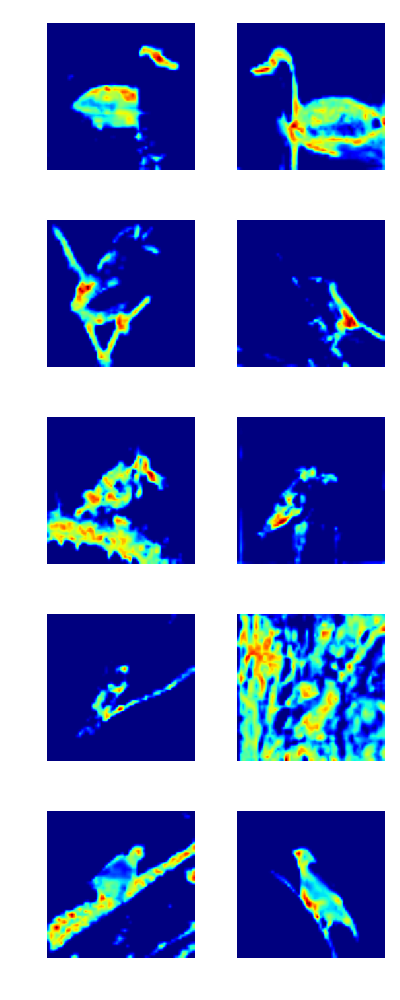}   
\end{minipage}} %
\subfigure[]{
\begin{minipage}{0.3\linewidth}
\centering
\includegraphics[scale=0.3]{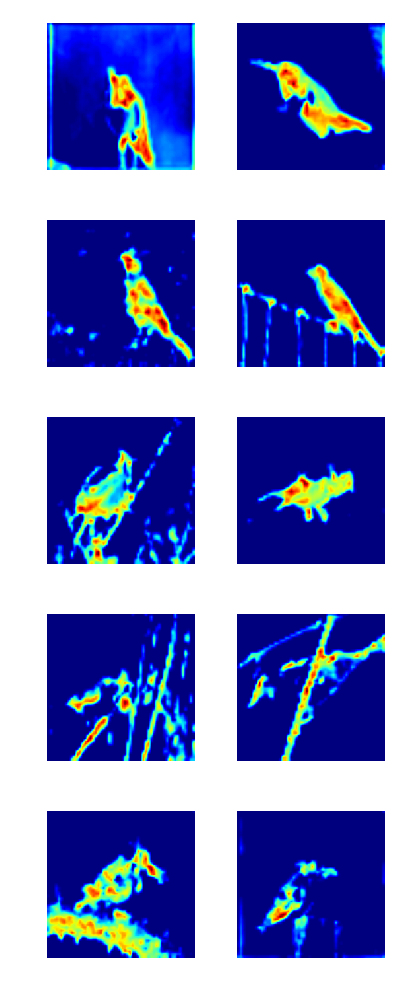}
\end{minipage}}
\caption{total information heat map}                                   \label{fig:total_heat_map}
\end{figure*}

\begin{figure*}[htbp]\centering                  
\subfigure[]{                    
\begin{minipage}{0.3\linewidth}\centering                                                          
\includegraphics[scale=0.3]{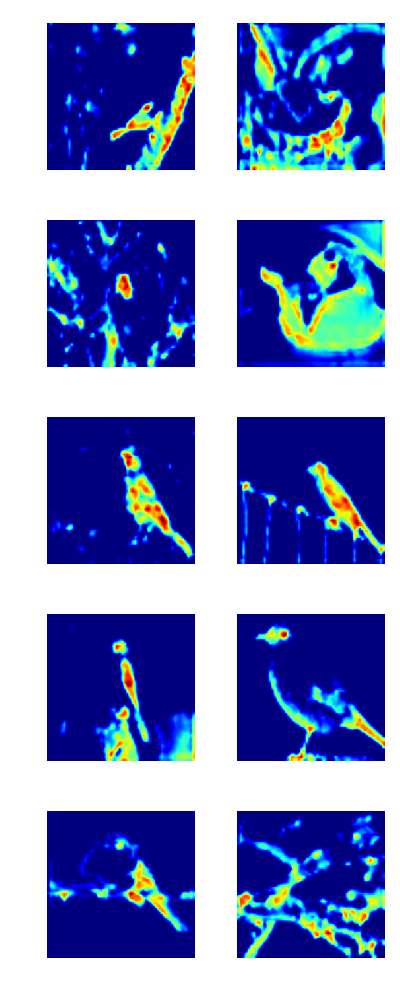}               
\end{minipage}}
\subfigure[]{                    
\begin{minipage}{0.3\linewidth}\centering                 
\includegraphics[scale=0.3]{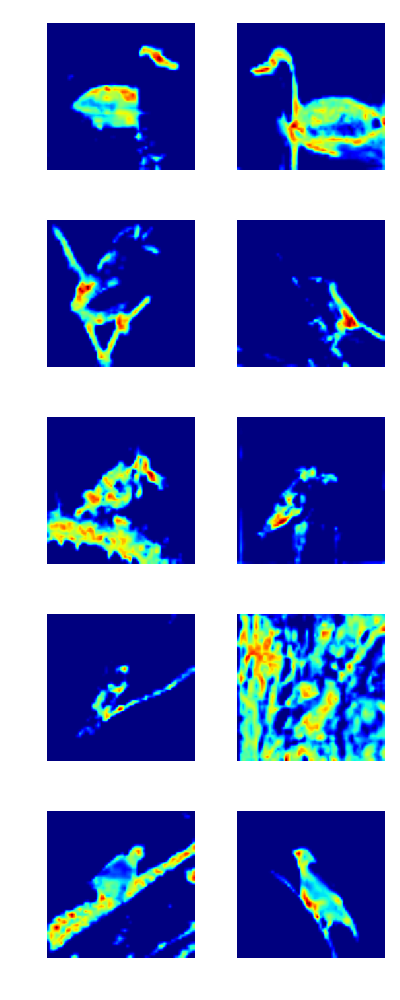}   
\end{minipage}} %
\subfigure[]{
\begin{minipage}{0.3\linewidth}
\centering
\includegraphics[scale=0.3]{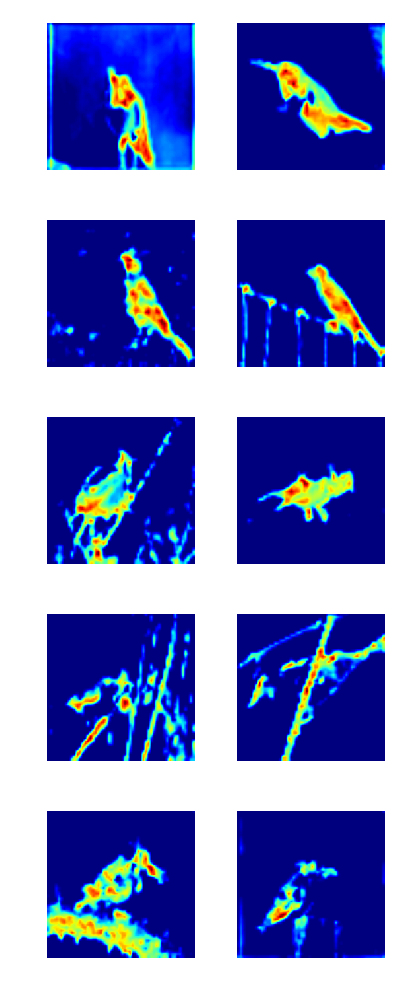}
\end{minipage}}
\caption{a mixed figure combining the total information heat map and the original image} %
\label{fig:total_mix}                                                        
\end{figure*}

\begin{figure*}[htbp]\centering                  
\subfigure[]{                    
\begin{minipage}{0.3\linewidth}\centering                                                          
\includegraphics[scale=0.3]{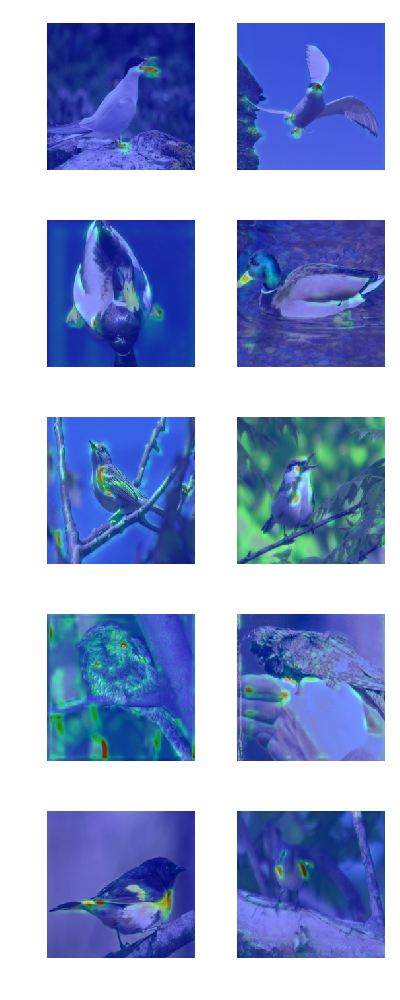}               
\end{minipage}}
\subfigure[]{                    
\begin{minipage}{0.3\linewidth}\centering                 
\includegraphics[scale=0.3]{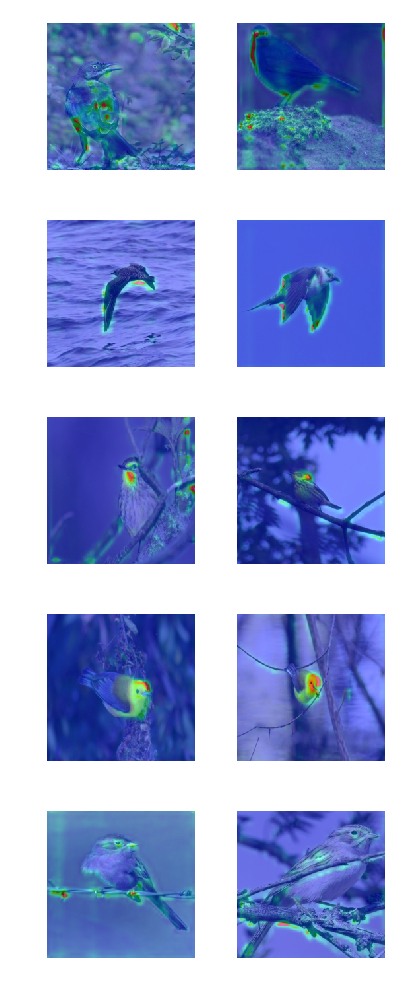}   
\end{minipage}} %
\subfigure[]{
\begin{minipage}{0.3\linewidth}
\centering
\includegraphics[scale=0.3]{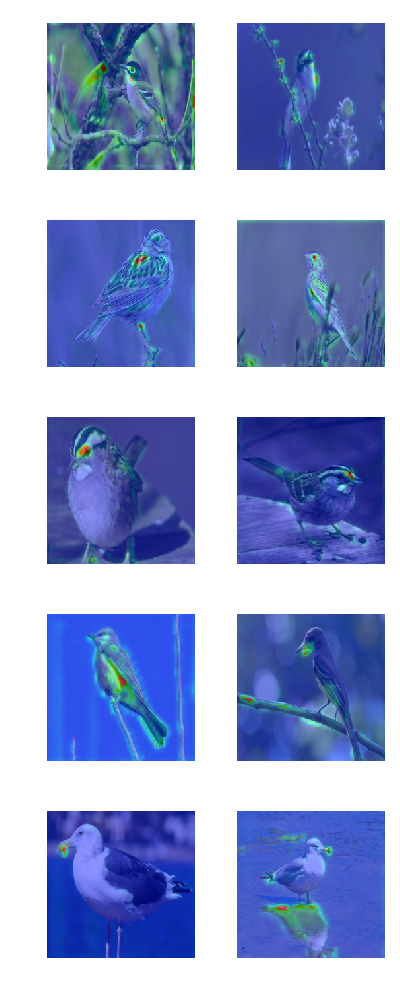}
\end{minipage}}
\caption{mixed figure combining the decision-related information heat map and the original image} %
\label{fig:decision_mix}                                                        
\end{figure*}

To further distinguish between total information and decision-related information, we visualize representations from the CUB-200-2011 dataset using a prototype network, which yields compelling visualization results. We treat the output of the prototype network’s Block 1 as local representations and the output of the average pooling (AvgPool) layer as global representations. Figures~\ref{fig:cub_images}, \ref{fig:total_heat_map}, \ref{fig:total_mix}, and \ref{fig:decision_mix} illustrate the heatmaps for total information and decision-related information, revealing clear and significant differences between the two.

\subsubsection{Few-Shot Learning}
The prototype network is a well-established method in few-shot learning, where the goal is to classify a query image into the correct category based on a support set consisting of a few labeled examples. This task is typically formulated as an \( N \)-way \( K \)-shot problem, where \( N \) represents the number of categories and \( K \) denotes the number of samples per category. For example, in a 5-way 1-shot scenario, the support set contains 5 categories, each with a single image. The prototype network classifies query images by comparing them to class prototypes derived from the support set.

\begin{figure*}[htbp]\centering                  
\subfigure[]{                    
\begin{minipage}{0.3\linewidth}\centering                                                          
\includegraphics[scale=0.3]{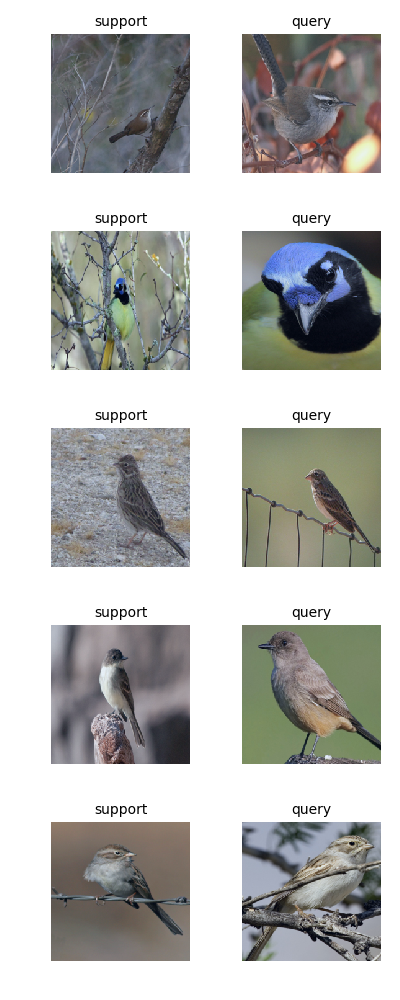}               
\end{minipage}}
\subfigure[]{                    
\begin{minipage}{0.3\linewidth}\centering                 
\includegraphics[scale=0.3]{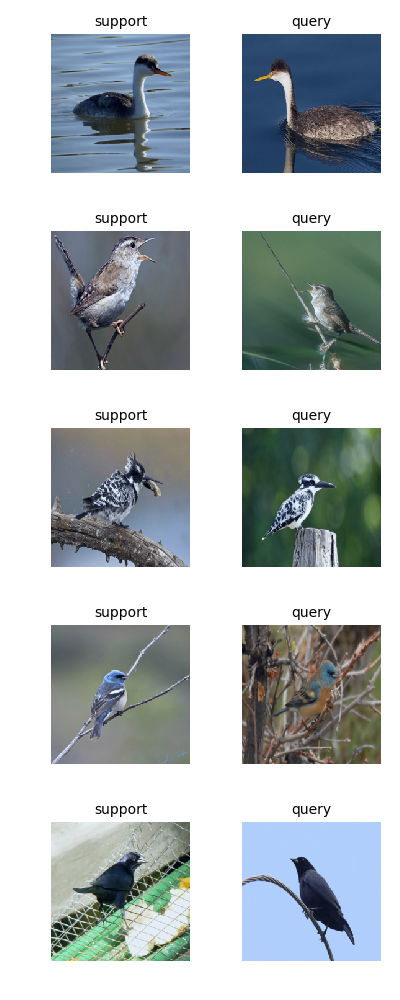}   
\end{minipage}} %
\subfigure[]{
\begin{minipage}{0.3\linewidth}
\centering
\includegraphics[scale=0.3]{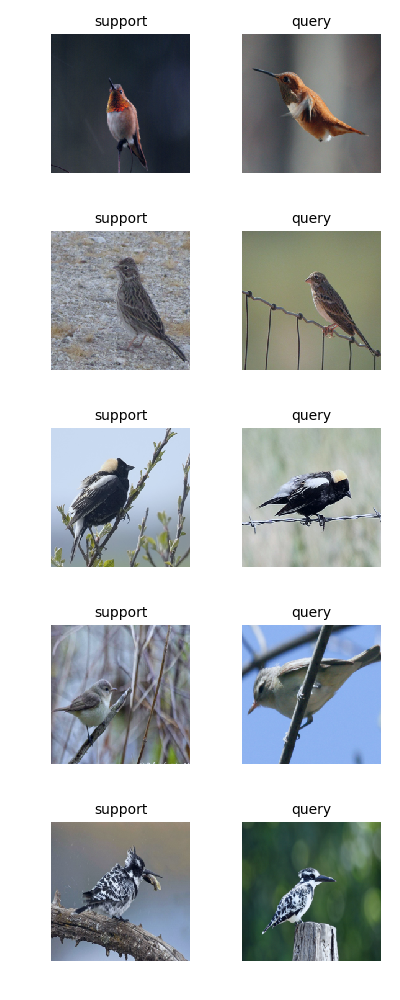}
\end{minipage}}
\caption{ images} %
\label{fig:cub_images}                                                        
\end{figure*}

For effective visualization, we adopt a 5-way 1-shot setting: the support set includes 5 categories with one image each, and we select 5 query images, one from each category in the support set. Figure~\ref{fig:cub_images} provides an example, where each column is labeled as either “support” or “query” to distinguish whether the 5 samples belong to the support set or the query set, respectively.


\section{Conclusion}
 We propose a mutual information (MI)-based framework that decomposes the relationship between local inputs and representations into three exhaustive components: total mutual information \( I(Z,X_i) \), decision-related information \( I(Z, X_i') \), and redundant information \( R(Z, X_i) = I(Z, X_i) - I(Z, X_i') \). This theoretically complete framework fully explains \( I(X, Z) \), surpassing partial methods like Grad-CAM by capturing all encoded content, decision drivers, and discardable noise. Using two lightweight modules, we visualize these components on ResNet and prototype network, confirming their interpretive power. This holistic approach establishes a robust tool for representation analysis.

\bibliography{main}
\bibliographystyle{icml2021}

\end{document}